\documentclass{article} 
\usepackage{hyperref}
\usepackage{url}

\usepackage[utf8]{inputenc}
\usepackage{amsmath,amssymb,latexsym,stmaryrd,graphicx} 
\newcommand{\ignore}[1]{}

\date{}

\usepackage{natbib}
\usepackage{graphicx}
\DeclareMathOperator*{\argmax}{argmax}
\DeclareMathOperator*{\argmin}{argmin}

\title{Information Theoretic Co-Training}

\author{David McAllester \\ TTI-Chicago}

\begin{document}

\begin{centering}
  \maketitle
\end{centering}

\begin{abstract}

  This paper introduces an information theoretic co-training objective for unsupervised learning.  We consider the problem of predicting the future.
  Rather than predict future sensations (image pixels or sound waves) we predict ``hypotheses'' to be confirmed by future sensations. More formally,
  we assume a population distribution on pairs $(x,y)$ where we can think of $x$ as a past sensation and $y$ as a future sensation.
  We train both a predictor model $P_\Phi(z|x)$ and a confirmation model $P_\Psi(z|y)$ where we view $z$ as hypotheses (when predicted) or facts (when confirmed).
  For a population distribution on pairs $(x,y)$ we focus on the problem of measuring the mutual information between $x$ and $y$.
  By the data processing inequality this mutual information is at least as large as
  the mutual information between $x$ and $z$ under the distribution on triples $(x,z,y)$ defined by the confirmation model $P_\Psi(z|y)$.
  The information theoretic training objective for $P_\Phi(z|x)$ and $P_\Psi(z|y)$ can be viewed as a form of co-training
  where we want the prediction from $x$ to match the confirmation from $y$.
  Initial experiments applying information theoretic co-training to unsupervised learning of phonetics are presented.
\end{abstract}

\section{Intuition and Formulation}

We Consider the problem of predicting the future from the past.  Intuitively
we are not interested in predicting raw future sense data such as image pixels.  Rather we are interested in predicting facts about the future as will be inferred from future sensations.
Here we consider the joint problem of (1) learning to convert sensation to
facts, and (2) learning to predict future facts. Information theoretic co-training aims to measure the
mutual information between past sensation and the future sensation by demonstrating the ability to predict future facts.

We formulate information theoretic co-training by letting ${\cal X}$ be a space of possible past sensations, ${\cal Y}$ be a space of possible future sensations,
and ${\cal Z}$ be a space of facts.  We assume a population
distribution ${\cal P}$ on sensation pairs $(x,y) \in {\cal X}\times {\cal Y}$.
We assume models $P_\Phi(z|x)$ and $P_\Psi(z|x)$ to predict future facts
from past and future sensations respectively.  Here $\Phi$ and $\Psi$ are parameter vectors and we assume that the probabilities are differentiable
in the parameters.

We assume that $z$ is most accurately estimated from $y$.
We then define a distribution on triples $(x,z,y)$ where $(x,y)$ is drawn from the population distribution ${\cal P}$ and $z$ is drawn from $P_\Psi(z|y)$.
We note that by the data processing inequality we have
$$I(x,y) \geq I_\Psi(x,z) = H_\Psi(z) - H_\Psi(z|x).$$
Here $H_\Psi(z)$ and $H_\Psi(z|x)$ are determined by the distribution on triples which is itself determined by $\Psi$.

The information theoretic co-training object takes into account the difficulty of empirically measuring the entropies $H_\Psi(z)$ and $H_\Psi(z|x)$.
In the phonetics experiment we have that $z$ has only 64 possible values which implies
that the entropy $H_\Psi(z)$ is at most six bits.  In this case $H_\Psi(z)$ can be approximated directly by the empirical marginal over $z$ in a large minibatch.
The entropy $H_\Psi(z|x)$ cannot in general be measured directly.  We assume that we can sample $(x,y)$ from the population, and sample $z$ from $P_\Psi(z|y)$,
but have no way computing $P_\Psi(z|x)$. However, following standard practice we can upper bound the entropy $H_\Psi(z|x)$ by the cross-entropy $H^{+}_{\Psi,\Phi}(z|x)$
where the model probability $P_\Phi(z|x)$ is computable.
\begin{eqnarray}
  I_\Psi(x,y) & \geq & H_\Psi(z) - H^{+}_{\Psi,\Phi}(z|x) \nonumber \\
  H^{+}_{\Psi,\Phi}(z|x) & = & E_{(x,y) \sim {\cal P},\;z \sim P_\Psi(z|y)} \;- \ln P_\Phi(z|x) \nonumber \\
  & = & H_\Psi(z|x) + KL(P_\Psi(z|y),P_\Phi(z|x)) \nonumber \\
    & \geq & H_\Psi(z|x) \nonumber
\end{eqnarray}
A first information theoretic co-training objective is then defined by
\begin{equation}
  \label{objective}
  \Psi^*\Phi^* = \argmax_{\Psi,\Phi} \; H_\Psi(z) - H^{+}_{\Psi,\Phi}(z|x).
\end{equation}
It is perhaps useful to rewrite (\ref{objective}) as
\begin{eqnarray}
  \Psi^* & = & \argmax_\Psi H_\Psi(z) - \left(\min_\Phi\;H^{+}_{\Psi,\Phi}(z|x)\right) \nonumber \\
  \Phi^* & = & \argmin_\Phi H^+_{\Psi^*,\Phi}(z|x) \nonumber \\
  \label{objectivea}
    & = & \argmin_\Phi E_{(x,y) \sim {\cal P},\;z \sim P_{\Psi^*}(z|y)} -\ln P_\Phi(z|x).
\end{eqnarray}
It should be noted that the objective (\ref{objectivea}) is the standard objective for training on labeled data.
In (\ref{objectivea}) $z$ replaces $y$ as a label for $x$.
The term $H_\Psi(z)$ in objective (\ref{objective}) encourages $\Psi$ to extract as much factual information from the future sensation as possible
while still making the extracted factual information predictable from the past.  Here $\Phi$ and $\Psi$  cooperate
to find agreement on a ``language'' (a semantics for symbols) grounded in sensation.

If $z$ is allowed to be a structured object, such as a sequence of symbols, then
$H_\Psi(z)$ becomes difficult to measure.  However, again following stadanrd practice, we can bound $H_\Psi(z)$ by a 
cross-entropy $H^+_{\Psi,\Theta}(z)$.
We then have the information theoretic co-training objective
\begin{eqnarray}
  \label{adversarial}
  \Psi^* & = & \argmax_\Psi \; \left(\min_\Theta H^+_{\Psi,\Theta}(z)\right) - \left( \min_\Phi\;H^+_{\Psi,\Phi}(z|x)\right) \\
  H^+_{\Psi,\Theta}(z) & = & E_{(x,y) \sim {\cal P}}\;\;E_{z \sim P_\Psi(z|y)} \;- \ln P_\Theta(z).\nonumber
\end{eqnarray}
Here $\Theta$ is adversarial to $\Psi$ and $\Phi$. Even in the case where $z$ is a structured object such as a string, it may be useful in practice to
bound the amount of information in $z$ by, for example, bounding the size of the alphabet and the length of the string.  This will make $H_\Psi(z)$
and $H_\Psi(z|x)$ smaller which should improve the numerical stability of the measured difference $H^+_{\Psi,\Theta^*}(z) - H^+_{\Psi,\Phi^*}(z|x)$.

\section{Related Learning Models}

{\bf Co-Training.} Information theoretic co-training is closely related to classical co-training (\cite{cotraining,PAC-co-training}).
Classical co-training assumes the same three spaces ${\cal X}$, ${\cal Y}$ and ${\cal Z}$
but takes the population ${\cal P}$ to be a distribution on triples $(x,z,y)$ where $z$ is not observed in the training data. The goal is to learn rules for predicting $z$
by training on the pairs $(x,y)$.  For this to be possible we need additional assumptions such as that $x$ and $y$ are independent given $z$ (in the population)
and that $H_{\cal P}(z)$ is large.  In information theoretic co-training, on the other hand, the population is assumed to be a distribution on $(x,y)$ only
and the goal is to measure the mutual information between $x$ and $y$.

Although the assumptions and theoretical analyses are different, the learning algorithms of information theoretic co-training and classical co-training are very similar.
The goal in classical co-training is to find hard (non-stochastic) classifiers $f:{\cal X} \rightarrow {\cal Z}$ and $g:{\cal Y} \rightarrow {\cal Z}$ so
as to maximize the probability over the draw of
$(x,y)$ that $f(x) = h(y)$ and, at the same time, to require that the values of $f(x)$ and $g(x)$ are diverse.
Information theoretic co-training makes the classifiers soft and makes the training objective information theoretic.

{\bf The Information Bottleneck.} Like information theoretic co-training, Tishby's information bottleneck \cite{bottleneck} assumes the spaces ${\cal X}$, ${\cal Y}$ and ${\cal Z}$ and assumes
a population distribution on the pairs $(x,y)$.  The objective is to train a model $P_\Psi(z|y)$ defining a distribution on triples $(x,z,y)$ using
the training objective
\begin{equation}
  \Psi^* = \argmax_\Psi\;I_\Psi(z,x) - \beta I_\Psi(z,y).
\end{equation}
In information theoretic co-training the second term is dropped and we retain only $I_\Psi(z,x)$.  One might immediately object that
the choice of $z=y$ maximizes $I_\Psi(z,x)$ so the objective is trivial if we drop the second term.  But the goal of information theoretic co-training is not
to maximize mutual information but rather to measure it.  Note that setting $z = y$ eliminates $\Psi$ from the information theoretic
co-training objective and we are left with setting $\Phi$
so as to minimize $H^+_{{\cal P},\Phi}(y|x)$.  This is the standard training objective for labeled data where we treat $y$ as a label.
This can also be viewed as conditional density estimation.  Conditional density estimation must be addressed to measure mutual information.
Setting $z=y$ is expected to yield a poor measurement of mutual information for two somewhat related reasons.  First,
the probabilistic modeling of raw sense data is difficult. Second $H^+_{{\cal P},\Theta}(y)$ and $H^+_{{\cal P},\Phi}(y|x)$ are both typically much larger than
$H^+_{\Psi,\Theta}(z)$ and $H^+_{\Psi,\Phi}(z|x)$.  So taking $z = y$ exposes one to numerical instability in taking the difference $H^+_{{\cal P},\Theta}(y) - H^+_{{\cal P},\Phi}(y|x)$.

{\bf Density Estimation.} Many approaches to unsupervised learning can be viewed as some form of density estimation.
Density estimation is the problem of modeling a probability distribution given the ability to draw samples. A paradigmatic example is language modeling.
In general we assume a population distribution ${\cal P}$ over some set ${\cal Y}$ and a model $P_\Psi(y)$ assigning a probability to each $y \in {\cal Y}$.
The density estimation objective is
\begin{eqnarray}
  \label{density}
  \Psi^* & = & \argmin_\Psi\;H^{+}_{{\cal P},\Psi}(y)  \\
  H^+_{{\cal P},\Psi}(y) & = & E_{y \sim {\cal P}}\;-\ln P_\Psi(y) \nonumber
\end{eqnarray}
The cross-entropy $H^{+}_{{\cal P},\Psi}(y)$ is an upper bound on the unknown, and typically unknowable, true entropy $H_{\cal P}(y)$.

Expectation maximization (EM) (\cite{EM}) and variational autoecoders (VAEs) (\cite{VAE}) optimize (\ref{density}) for the case where
$P_\Psi(y)$ is a marginal distribution over a latent variable $z$.
\begin{equation}
  \label{marginal}
  P_\Psi(y) = \sum_z\; P_\Psi(z,y)
\end{equation}
Here $P_\Psi(z,y)$ is typically a generative model where $y$ is generatively derived from $z$.

Data compression algorithms also implicity optimize (\ref{density}).  By Shannon's source coding theorem the most efficient code for instances drawn
from a given population uses a number of bits equal to the entropy of the population distribution.
The training objective (\ref{density}) can be interpreted as optimizing the compressed bits per sample when drawing from the population but coding for
the model.

Information theoretic co-training as defined by (\ref{objective}) and (\ref{adversarial}) differs from density estimation as defined by (\ref{density})
in that information theoretic co-training
uses only probability models for the ``facts'' $z$ --- in information theoretic co-training there is no attempt to model distributions on the sensations.

{\bf GANs.} Generative adversarial networks (GANs) (\cite{Schmidhuber-GANs,GANs}) are similar to variational autoencoders in that
they define a generative model $P_\Psi(z,y)$ where $y$ is generated from $z$ and where we are interested in the  marginal distribution (\ref{marginal}).
However, in GANs
there is no attempt to optimize, or even measure, a cross-entropy (\ref{density}).  Instead one define a distribution $Q_\Psi$ on pairs $(y,\ell)$
by drawing $y$ with equal probability either from the population
distribution ${\cal P}$ or the model distribution $P_\Psi$ and setting $\ell = 1$ is $y$ is drawn from ${\cal P}$ and $\ell = -1$ if $y$ is drawn from $P_\Psi$.
A discriminator model $P_\Phi(\ell|y)$ must predict which distribution $y$ was drawn from.
The GAN objective is
\begin{eqnarray}
  \label{GAN}
  \Psi^* & = & \argmax_\Psi\; \min_\Theta \; H^+_{\Psi,\Theta}(\ell|y) \\
  H^+_{\Psi,\Theta}(\ell|y) & = & E_{(y,\ell) \sim Q_\Psi}\;-\ln P_\Theta(\ell|y) \nonumber
\end{eqnarray}
In (\ref{GAN}) the generator $\Psi$ is trying to generate values such that discriminator $\Theta$ cannot distinguish values generated from $P_\Psi$
from values drawn from ${\cal P}$.
InfoGANS (\cite{InfoGANS}) add a term to the GAN objective to increase the mutual information between certain components of the latent variable $z$ and the generated variable $y$.
This encourages the model to use independent components of the information in $z$.

A major issue with GANs is the lack of an objective
measure of performance.  The ability to fool a particular discriminator architecture does not imply low cross-entropy as defined by (\ref{density}). It is quite
plausible that large modes of the population density are omitted from the generator distribution (the problem of mode dropping).
In contrast, information theoretic co-training provides a quantitative performance measure.

\section{Unsupervised Learning of Phonetics}

\begin{figure}
  \centerline{\includegraphics[width= 3.5in]{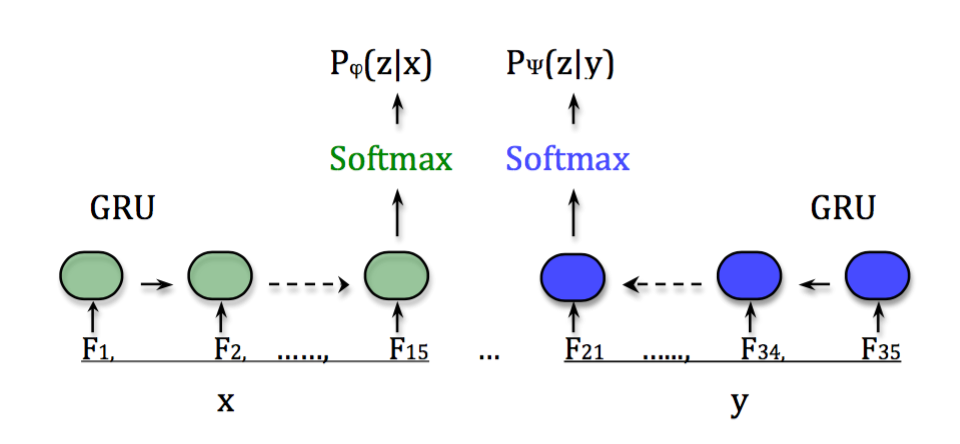}}
  \caption{{\bf The architecture.} A sliding window of 35 speech frames is slid over an utterance. At each placement of the window the architecture computes
    $P_\Psi(z|y)$ and $P_\Phi(z|x)$. Here $z$ ranges over an alphabet of 64 unsupervised symbols.
    The GRUs use bias vectors in their weight layers and have an initial hidden state vector parameter. The GRUs use 64 dimensional hidden states.  Each utterance is run as a single minibatch
    and the loss function for the minibatch is $H^+_{\Psi,\Phi}(z|x) - H_\Psi(z|u)$ where $H_\Psi(z|u)$ is the entropy of the empirical average of
    $P_\Psi(z|y)$ over that minimibatch (the utterance $u$).  Conditioning on the utterance avoids the possibility of using speaker identification as a means of achieving co-training
    agreement. To get the correct expectation the first term in the loss should be averaged over the minibatch
    while the second term, which is already based on an average distribution, should not.}
  \label{architecture}
\end{figure}

We performed information theoretic co-training on the TIMIT training data (\cite{TIMIT}).
The TIMIT training data consists of 3.7 thousand utterances with an average of 304 speech frames per utterance yielding
1.1 million frames of speech. Each frame is labeled with a 39 dimensional MFCC feature vector. We used the normalized data where each utterance is normalized
so that the mean over the utterance of the squared norm of the vectors is 1.
The TIMIT training data is converted to a set of pairs $(x,y)$
by passing a 35 frame window over the data where $x$ is taken to be the first 15 frames and $y$ is taken to be the last 15 frames so that $x$ and $y$ are separated by five frames.
We take ${\cal Z}$ to be an alphabet of 64 symbols.  The models $P_\Psi(z|y)$ and $P_\Phi(z|x)$ are computed by GRUs (\cite{GRU,LSTM}).
The architecture is shown in figure~\ref{architecture}. More details are given in the figure caption.

\begin{figure}
  \centerline{\includegraphics[width= 3.5in]{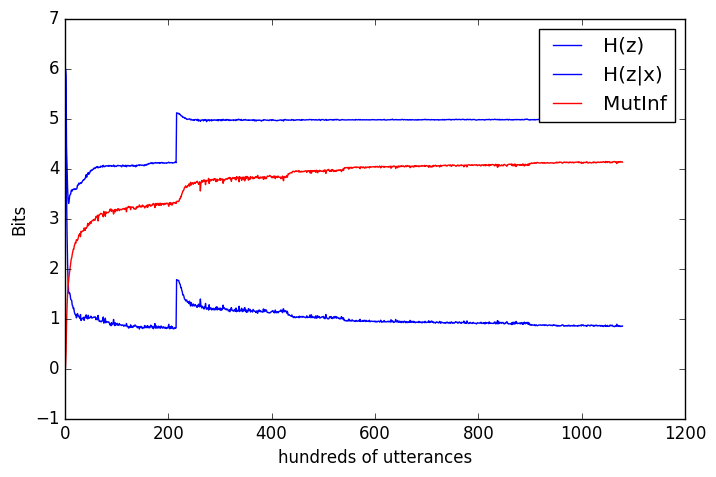}\hspace{1em}}
  \caption{{\bf The learning curve.}  The model was trained with vanilla SGD. The learning rate was initialized to .4 for 216 hundred utterenace (six epochs).  At that point the surviving 21 symbols were
    ``cloned'' - for each of the 21 live symbols $z$  one of the dead symbols was replaced with a noisy version of $z$.  The cloning event adds one bit
    of information to both $H(z)$ and $H(z|x)$.  After another 216 hundred utterances (six more epochs), the learning rate was reduced to .2; at 540 hundred utterances the learning
    rate was dropped to .1; at 900 hundred utterances the learning rate was dropped to .05; and the training was stopped at 1080 hundred utterances (30 epochs).
    After unsupervised training the majority symbol predicted from $x$
    agreed with the majority symbol predicted from $y$ on 78.6\% of the window placements on the development set. $H_\Psi(z|u)$ on the development set was 5.0 bits
    (a perplexity of 32).}
  \label{learningcurve}
\end{figure}

The model was trained with vanilla SGD. In the first few epochs all but 21 of the 64 available symbols died ---
they were assigned essentially zero probability by both $P_\Psi(z|y)$ and $P_\Phi(z|x)$ for all placements of the window.
After six epochs the surviving 21 symbols were ``cloned'' --- a noisy copy of each surviving symbol was created yielding 42 surviving symbols.
The model was then trained for another 24 epochs (totaling 30 epochs).  The learning curve is shown in figure~\ref{learningcurve}.
More details are given in the figure caption.

\begin{figure}
  \centerline{\includegraphics[width= 3.5in]{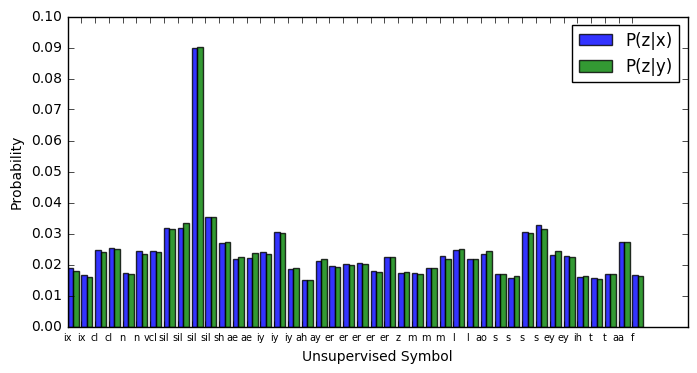}\hspace{1em}}
  \caption{{\bf The distribution of the unsupervised symbols.} The bar chart shows the distribution of the unsupervised symbols as measured by averaging
    $P_\Psi(z|x)$ (blue) and averaging $P_\Phi(z|y)$ (green) over the development data.
    Each unsupervised symbol is tagged with its most likely phoneme.  Note that four
    unsupervised symbols are tagged with silence ('sil') and four with 's'. Only 21 phonemes occur as the tag of a symbol.
    Except for one of the silence symbols, the distribution is fairly
    uniform as one might expect from maximizing $H^+_{{\cal P},\Psi}(z|u)$.}
  \label{frequency}
\end{figure}

After training the model was used to label each frame of the data with an unsupervised symbol.  More specifically, for each frame of data a window is placed centered on that frame
and the frame is labeled with the symbol $z$ maximizing $P_\Psi(z|y)$.  After labeling the frames with unsupervised symbols, each unsupervised symbol $z$ is tagged with the
majority phoneme label for the frames labeled with $z$.
Although 42 unsuervised symbols survived the training, many of these symbols get tagged with the same phoneme.  Four symbols are tagged with silence ('sil')
and four with 's'.  Only 21 phonemes get used as tags. The distribution over unsupervised symbols is shown in figure~\ref{frequency}.

\begin{figure}
  \centerline{\includegraphics[width= 3.5in]{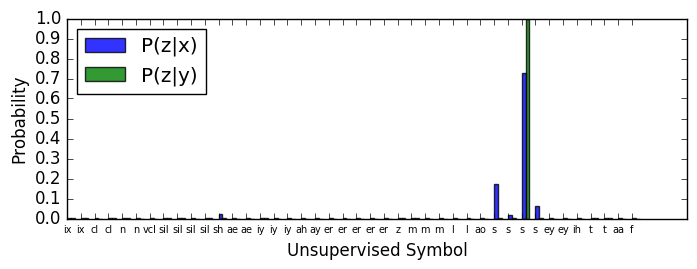}\hspace{1em}}
  \caption{{\bf Probabilities at a typical frame.} Note that $P_\Psi(z|y)$ is essentially deterministic while $P_\Phi(z|x)$ represents a more reasonably uncertain belief.
    This asymmetry is due to the asymmetry in the cross entroy $H^+_{\Psi,\Phi}(z|x)$. This is discussed from a game-theoretic perspective in the text.
    The most likely $z$ under $P_\Psi(z|y)$ agrees with the most likely $z$ under $P_\Phi(z|x)$ in 78.6\% of frames.}
  \label{frame}
\end{figure}

Figure~\ref{frame} shows $P_\Psi(z|y)$ and $P_\Phi(z|x)$ for a typical placement of the model window.
We get that $P_\Psi(z|y)$ is sharp, essentially deterministic, while $P_\Phi(z|x)$ represents a more reasonably uncertain posterior belief.  This asymmetry is due to
the asymmetry of the cross-entropy
$$H^+_{\Psi,\Phi}(z|x) = E_{z \sim P_\Psi(z|y)} - \ln P_\Phi(z|x).$$
This has an intuitive explanation if we think of $\Psi$ and $\Phi$ as agents in a cooperative game.
Suppose that $\Psi$ has a reasonably uncertain believe about $\Phi$'s prediction of $z$.
In this case $\Psi$'s best strategy is to place the bet deterministically on the value of $z$ for which the expectation of $- \ln P_\Phi(z|x)$ is largest.
This is not true of $\Phi$ --- if $\Phi$ makes $P_\Phi(z|x)$ deterministic
then there a large probability of an infinite loss if $\Psi$ guesses wrong.

\begin{figure}
   \centerline{\includegraphics[width= 4.5in]{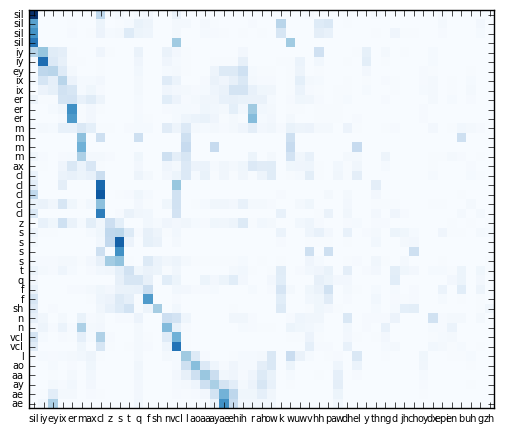}}
  \caption{{\bf The confusion matrix.}
    Each row of the confusion matrix gives the distribution over the actual phonetic label when a certain phonetic prediction is made.
    Only 21 of the phonemes are ever predicted. The accuracy on frames labeled with one of those 21 phonemes is 47.3\%.
    On all frames the accuracy is 35.1\%.  The most common phonetic label ('sil') occurs in 13/\% of the frames. State of the art
    supervised methods achieve approximately 80\% frame level accuracy .  We were unable to find comparable unsupervised results.}
  \label{confusion}
\end{figure}

Figure~\ref{confusion} gives a confusion matrix. This is computed by first labeling each frame with an unsupervised symbol by placing a window centered at that frame and selecting the
unsupervised symbol with largest probability under $P_\Psi(z|y)$.  Labeled data is then used to map each unsupervised symbol to the majority phoneme at the frames labeled with that symbol
yielding a predicted phoneme at each frame. The resulting predictions turn out to be limited to 21 phonemes.
If we consider only frames with gold labels from this set of 21, the model achieves an accuracy of 47.3\%.  On all frames the accuracy is 35.1\%.
The most common phonetic label ('sil') occurs in 13/\% of the frames. State of the art
supervised methods achieve approximately 80\% frame level accuracy .  We were unable to find comparable unsupervised results.
It should be kept in mind that this is a very first attempt.

\section{Conclusions}

Information theory already plays a central role in the training objectives typically used in deep learning.  Information theoretic co-training
introduces a novel information theoretic objective for unsupervised learning in which one can avoid any attempt to measure the entropy, or conditional entropy,
of raw sense data. Information theoretic co-training can also be viewed as a way of measuring mutual information by developing a ``language'' for carrying that information
where the entropy of the facts stated in that language is small compared to the entropy of raw sense data.

The experiments presented here are very preliminary. We expect that much stronger results in unsupervised learning of phonetics are possible with just a little more
experimentation.  We are also anxious to experiment with other data such as pairs of images from video and natural language translation pairs.


\end{document}